# SupWMA: Consistent and Efficient Tractography Parcellation of Superficial White Matter with Deep Learning

*Tengfei Xue*[1,2], *Fan Zhang*[1]\*, *Chaoyi Zhang*[2], *Yuqian Chen*[1,2], *Yang Song*[3], *Nikos Makris*[1,4],
*Yogesh Rathi*[1], *Weidong Cai*[2], *Lauren J. O'Donnell*[1]

[1] Brigham and Women's Hospital, Harvard Medical School, USA
[2] School of Computer Science, University of Sydney, Australia
[3] School of Computer Science and Engineering, University of New South Wales, Australia
[4] Center for Morphometric Analysis, Massachusetts General Hospital, Boston MA

## ABSTRACT

White matter parcellation classifies tractography streamlines into clusters or anatomically meaningful tracts to enable quantification and visualization. Most parcellation methods focus on the deep white matter (DWM), while fewer methods address the superficial white matter (SWM) due to its complexity. We propose a deep-learning-based framework, *Superficial White Matter Analysis (SupWMA)*, that performs an efficient and consistent parcellation of 198 SWM clusters from whole-brain tractography. A point-cloud-based network is modified for our SWM parcellation task, and supervised contrastive learning enables more discriminative representations between plausible streamlines and outliers. We perform evaluation on a large tractography dataset with ground truth labels and on three independently acquired testing datasets from individuals across ages and health conditions. Compared to several state-of-the-art methods, SupWMA obtains a highly consistent and accurate SWM parcellation result. In addition, the computational speed of SupWMA is much faster than other methods.

***Index Terms***— diffusion MRI, tractography, superficial white matter parcellation, deep learning, point cloud

## 1. INTRODUCTION

Diffusion magnetic resonance imaging (dMRI) tractography is the only non-invasive method to map brain white matter connections [1]. Performing whole-brain tractography generates a tractogram of the entire white matter, covering the deep white matter (DWM) that connects distant cortical regions [2], and the superficial white matter (SWM) that includes short-range association connections (u-fibers) connecting adjacent and nearby gyri [3]. A tractogram can contain hundreds of thousands of streamlines, which are not directly useful to clinicians and researchers for quantification or visualization. Therefore, tractography parcellation is needed. While most tractography parcellation methods currently focus on the DWM [4–8], few methods can parcellate the SWM [3,9–11].

The existing SWM tractography parcellation methods use either region of interest (ROI)-based selection or streamline clustering. ROI-based methods parcellate tractography based on the ROIs streamlines end in and/or pass through [9,12]. These ROI-based methods are commonly used but highly depend on the ROI parcellation scheme. Streamline clustering methods group streamlines based on the similarity of their geometric trajectories [10,11,13]. Such streamline clustering methods for SWM parcellation are automatic and can leverage curated SWM atlases, but challenges remain to achieve consistent parcellation across subjects and reduce runtime.

In recent years, deep-learning-based methods [6–8,14,15] have been successful for fast and consistent tractography parcellation, but representing dMRI data in a way that can best take advantage of deep networks is still an open challenge. Voxel-based [6,16] methods take volumetric image data (e.g., fiber orientation distribution function (FOD) peaks [6]) and predict a tract's presence and/or orientation for each voxel. Streamline-based methods [7,15,17] encode streamlines into different features as the input of deep networks. For example, streamlines have been encoded as images to enable effective processing by Convolutional Neural Networks (CNNs). DCNN+CL+ATT [15] uses a 2D image containing streamline point spatial coordinates, and DeepWMA [7] represents streamlines as 3-channel 2D images. However, points are geometric primitives and an image representation is not straightforward. In addition, the ambiguity of streamline data (the points along a streamline can equivalently be represented in forward or reverse order) poses challenges when using an image representation in CNNs. Point clouds, as an important geometric data format [18], can potentially enable efficient and discriminative representations for streamlines. RAS (Right, Anterior, Superior) coordinates of streamline points can be the input for point-cloud-based deep networks, as used in [19] for tractography filtering (binary classification of streamlines). However, to our knowledge, no deep learning methods have focused on SWM parcellation, and point-cloud-based deep networks have not yet been used for white matter parcellation, in particular for SWM parcellation.

\*Correspondence: fzhang@bwh.harvard.edu

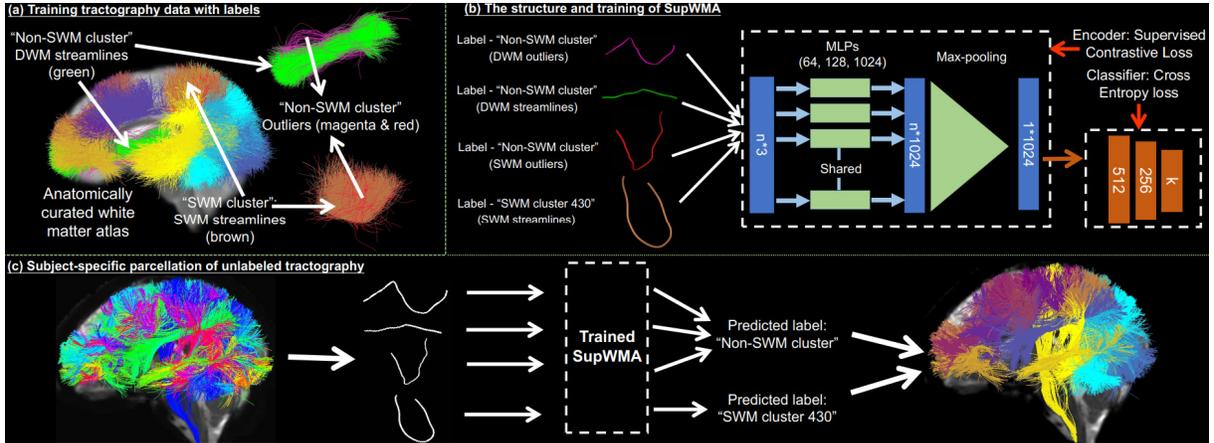

**Fig. 1**. Overview of the SupWMA framework: (a) samples of training tractography data, (b) deep network structure and parcellation model training, (c) parcellation of unseen testing datasets.

In this paper, we propose a novel deep learning framework, *Superficial White Matter Analysis (SupWMA)* for SWM parcellation from whole-brain tractography. SupWMA is designed based on a point-cloud-based network [20] with supervised contrastive learning [21] to classify streamlines into SWM clusters and remove outliers. Our contributions are as follows. 1) We propose a deep learning framework for the task of SWM parcellation, and we obtain fast and consistent results across populations and dMRI acquisitions. 2) We modify the point-cloud-based network structure to preserve streamline pose and orientation information (because location in the brain is important for classification). An advantage of our point-cloud-based network is that streamlines with equivalent forward and reverse point orders (e.g., from cortex to subcortical structures or vice versa) can have the same global shape feature representation. 3) We investigate supervised contrastive learning to obtain more discriminative representations between plausible streamlines and outliers by leveraging the supervised contrastive loss and the label information of the training dataset.

## 2. METHODOLOGY

### 2.1. dMRI Datasets and Tractography

A high-quality large-scale dataset with 1 million labeled streamlines was used for model training and validation. This dataset was derived from an anatomically curated white matter tractography atlas [11]. In brief, the atlas was created from 100 young healthy adults in the Human Connectome Project (HCP) [22] and was annotated by a neuroanatomist. The training data (Fig. 1(a)) includes labels for 198 expert-curated SWM clusters (206602 streamlines). For our SWM framework, we group all non-SWM streamlines (DWM and outlier streamlines) together using a label of "non-SWM cluster" (793398 streamlines).

For experiments, we used publicly available data (150 subjects) from three independently acquired datasets with different imaging protocols across ages and health conditions. (1) HCP dataset [22]: 100 young healthy adults[1] (age: 22 to 35y, 29.0±3.5), dMRI acquisition parameters: b = 3000s/mm$^2$, 108 directions, TE/TR = 89/5520ms, resolution = 1.25mm$^3$; (2) Adolescent Brain Cognitive Development (ABCD) [23] dataset: 25 teenagers (age: 9 to 11y, 10.1±0.7), dMRI acquisition parameters: b = 3000s/mm$^2$, 96 directions, TE/TR = 88/4100ms, resolution = 1.7mm$^3$; (3) Parkinson's Progression Markers Initiative (PPMI) [24] dataset: 25 elderly adults (age: 51 to 75y, 63.8±5.4), including Parkinson's disease (PD) patients and healthy individuals. dMRI acquisition parameters: b = 1000s/mm$^2$, 64 directions, TE/TR = 88/7600ms, resolution = 2mm$^3$. The two-tensor Unscented Kalman Filter (UKF)[2] method [25], as used in [11,26,27], was applied to generate whole-brain tractography for all subjects in all datasets above. Tractography was performed in 3D Slicer[3] software [28] via the SlicerDMRI[4] project [29,30].

### 2.2. Pointwise Encoding Structure

PointNet [20] is widely used for point cloud classification and segmentation [18]. It includes a shared multi-layer perceptron (MLP), a symmetric aggregation function, and fully connected (FC) layers, as well as data-dependent transformation nets. The transformation nets are designed to transform (affine) input point clouds into a canonical space [20]. However, the labeling of streamlines is sensitive to rotation and translation. Therefore, we remove the transformation nets from our framework to preserve significant information about the spatial position of streamlines in the brain. Also, the computational speed of the network is improved with transformation nets removed.

Our network can be divided into two parts: the encoder *Enc(·)* that extracts the global feature for each streamline and the classifier *Cla(·)* that predicts the streamline label. For *Enc(·)*, the input is the RAS coordinates of streamline points, denoted as $X=(x_1, x_2, ..., x_n)$, where $x_i$ is a 3-D coordinate vector of point *i*. Therefore, the dimension of input is n×3 (n = 15, as used in [7,11,31]). Each point $x_i$ is individually encoded by a shared MLP of three layers that have 64, 128, and 1024 output dimensions, respectively, generating an output *X'* with n×1024 dimensions. Then, a symmetric function (max-pooling [20]) aggregates the encoded features *X'* to form a 1024-dimension global shape feature *g* of the streamline. *g* is invariant to the order of points along a streamline, so that streamlines with equivalent forward and reverse point orderings are allowed to have the same global shape feature (representation). Finally, the *Cla(·)*, consisting of three FC layers with sizes of 512, 256, and *k* (number of output classes), is used for streamline class prediction.

---

[1] These subjects are different from those used in the atlas.
[2] github.com/pnlbwh/ukftractography
[3] www.slicer.org
[4] dmri.slicer.org

### 2.3. Supervised Contrastive Learning

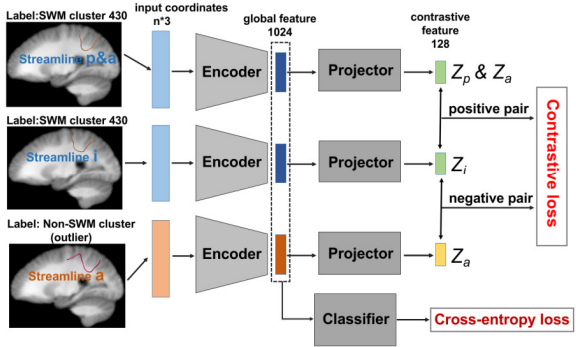

**Fig. 2**. The process of supervised contrastive learning and downstream task training

Supervised contrastive learning (SCL) [21] extends self-supervised contrastive learning [32] to a fully-supervised mode by proposing a supervised contrastive loss, which aims to pull global features (outputs of *Enc(·)*) with the same class label closer in the latent space and push apart global features with different class labels. In our task, as streamlines in the SWM outlier class can have a high similarity in geometry to those in the SWM clusters, SCL is used to assist *Enc(·)* in extracting more discriminative features for streamlines.

In SCL training for *Enc(·)* (Fig. 2), a projector head *Proj(·)* [21,32] is added with two additional FC layers of sizes 1024 and 128 followed by a normalization layer. Therefore, the contrastive feature *z* of input *X* for calculating contrastive loss is formed as $z=Proj(g)=Proj(Enc(X))$. *Proj(·)* may retain more instance-specific streamline information in the global feature *g*, benefiting downstream tasks [32].

The supervised contrastive loss is shown below:
$$\mathcal{L}_{out}^{sup} = \sum_{i \in I} \mathcal{L}_{out,i}^{sup} = \sum_{i \in I} \frac{-1}{|P(i)|} \sum_{p \in P(i)} \log \frac{\exp(z_i \cdot z_p / \tau)}{\sum_{a \in A(i)} \exp(z_i \cdot z_a / \tau)}$$
where *I* is the streamline set in a training batch ($i \in I \equiv \{1, ..., M\}$); *P(i)* is the streamline set that has the same class label as streamline *i* ($p \in P(i)$); *A(i)* is the set of all other streamlines in *I* except for streamline *i* ($a \in A(i) \equiv I \setminus \{i\}$); $z_i$, $z_p$ and $z_a$ are contrastive features obtained from *Proj(·)* for streamlines *i*, *p* and *a*; τ (temperature) is a pre-defined hyperparameter set to be 0.1 as suggested in [32].

### 2.4. Model Training and Testing

Fig. 1(b) gives an overview of our network, and Fig. 2 summarizes the training procedure that includes two phases: contrastive learning and downstream learning [21,32]. Hyperparameters are tuned and reported as follows. In contrastive learning phase, *Enc(·)* and *Proj(·)* are trained with supervised contrastive loss. The learning rate is 0.01, and the batch size is 6144 (as suggested in [21]). In downstream learning phase, the parameters of *Enc(·)* are frozen as in [21], and *Proj(·)* is untouched. *Cla(·)* takes *g* (the output of *Enc(·)*) as the input and is trained with cross-entropy loss for streamline classification. The learning rate is 0.001, and the batch size is 1024. Both training phases utilize Adam [33] as the optimizer with no weight decay. Training and validation were performed with Pytorch (v1.5) on a NVIDIA GeForce RTX 2080 Ti GPU machine.

Fig. 1(c) demonstrates parcellation of unlabeled tractography data (i.e., testing data). The subject-specific tractography data is first transformed into the atlas space by registering (affine) the b0 image of the subject to the mean T2w image of the atlas population using 3D Slicer. Then our trained model takes streamline features of the transformed tractography as input to predict the class label for each streamline, obtaining the classification result of 198 SWM clusters and one non-SWM cluster.

## 3. EXPERIMENTS AND RESULTS

### 3.1. Performance on Datasets with Ground Truth

We first evaluated our method on the dataset with ground truth streamline labels using 5-fold cross-validation. For experimental comparison, we included two deep-learning-based state-of-the-art (SOTA) tractography parcellation methods: DCNN+CL+ATT[5] [15] and DeepWMA[6] [7]. Both DCNN+CL+ATT and DeepWMA are designed for DWM parcellation using CNNs and streamline spatial coordinate features. In our study, we trained their networks and fine-tuned hyperparameters based on the suggested settings in their papers and released code. In addition to the comparison to SOTA, we also performed an ablation study: the original PointNet implementation (with transformation nets equipped), and SupWMA$_{no-SCL}$ that does not use supervised contrastive loss in the training compared to SupWMA.

For each of the aforementioned methods, we computed the accuracy and macro F1-score metrics, which have been widely used for tractography parcellation [7,15,17,34]. For each cross-validation fold, the overall accuracy of streamline classification is calculated, and the mean and standard deviation of macro F1-score across 199 streamline classes are also reported, as well as the average of the metrics across the five folds. We also included the floating point operations (FLOPs), which measure the number of required operations performed for model inference [20], for evaluating the efficiency of each method.

---

[5] github.com/HaotianMXu/Brain-fiber-classification-using-CNNs

[6] github.com/SlicerDMRI/DeepWMA

Table 1. Quantitative comparisons on ground truth dataset.

|  | Methods | Accuracy | F1-score | FLOPs / streamline |
|---|---|---|---|---|
| SOTA comparison | DeepWMA | 92.7% | 73.3±7.9% | 40.9M |
|  | DCNN+CL+ATT | 95.1% | 82.3±4.4% | 36.8M |
| Ablation study | PointNet | 96.1% | 86.5±3.5% | 9.6M |
|  | SupWMA$_{no\text{-}SCL}$ | 96.2% | 86.8±3.3% | 2.8M |
|  | SupWMA (ours) | **96.7%** | **88.5±3.0%** | **2.8M** |

Compared to the SOTA methods, SupWMA achieved the highest mean accuracy and macro F1-score with the lowest standard deviation. Also, the FLOPs of SupWMA are much lower than other SOTA methods. The ablation study shows that SupWMA$_{no\text{-}SCL}$ improves the model performance and reduces the FLOPs compared to the baseline PointNet method, because transformation nets are not helpful for the task of SWM parcellation and also increase the network size. SCL can help the *Enc(·)* extract more discriminative global features for streamlines. Therefore, it further enhances the results of accuracy and macro F1-score for SupWMA.

### 3.2. Performance on Datasets without Ground Truth

We then performed experiments on the three independently acquired datasets (HCP, ABCD, and PPMI) without ground truth to evaluate our method's ability to generalize to unseen tractography data. The three datasets are across populations from different ages (children, adults, older adults) and health conditions (healthy controls, Parkinson's patients). In addition to DCNN+CL+ATT and DeepWMA, we include another SOTA method, WMA[7] [11,31]. We note that WMA performs tractography parcellation by applying the same anatomically curated atlas (see [11] for details) as we used for generating our training data. For each method, we quantified the cluster identification rate (CIR), a metric that measures the success of white matter cluster identification in the absence of ground truth parcellation [7,11]. In our study, a cluster was considered to be successfully detected if there were at least 20 streamlines (as in [7,11]). In addition, we also provide a visualization of the identified SWM clusters in an example individual subject for each dataset.

Table 2 gives the comparison results using the CIR. Our method performs best on all three testing datasets compared to the three SOTA methods. Fig. 3 gives visualization results of example clusters for each testing dataset and method. All methods have relatively good performance on clusters shown in blue and cyan (except DeepWMA on HCP). However, SupWMA identifies more reasonable streamlines (magenta clusters) than other methods on the HCP and ABCD datasets.

Table 2. Cluster identification rates across methods for three testing datasets.

| Methods | HCP | ABCD | PPMI |
|---|---|---|---|
| WMA | 96.2±3.4% | 82.5±7.5% | 93.2±5.2% |
| DeepWMA | 97.0±2.9% | 82.9±5.7% | 93.8±3.7% |
| DCNN+CL+ATT | 97.9±2.5% | 83.8±6.0% | 93.8±4.0% |
| SupWMA (ours) | **98.5±2.1%** | **88.4±5.3%** | **94.6±4.1%** |

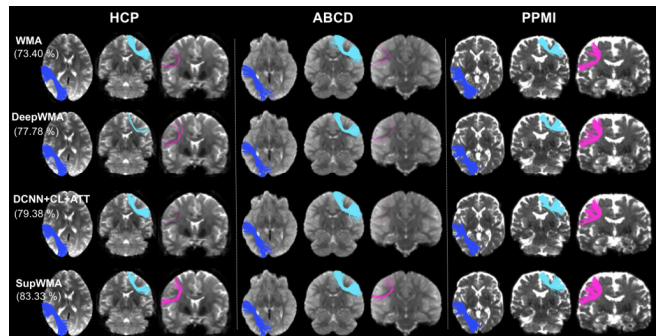

**Fig. 3**. Visualization of example individual clusters. For each testing dataset, the subject that has the lowest CIR is selected for visualization. The average CIR (across the three subjects) is displayed for each method.

Computation time is tested on a Linux workstation with CPU and GPU using a randomly selected subject (0.44 million streamlines). Our method has the shortest computation time in both testing environments using CPU only and using CPU and GPU. Also, benefiting from the efficient network structure, our method has the smallest increase of computation time when changing from CPU and GPU mode to CPU only mode.

Table 3. Comparisons of computation time across methods.

| Method | CPU only | CPU + GPU |
|---|---|---|
| WMA | 101min | --- |
| DeepWMA | 4min29s | 2min25s |
| DCNN+CL+ATT | 4min48s | 1min56s |
| SupWMA (ours) | **1min59s** | **1min22s** |

### 4. CONCLUSION

In this study, we proposed SupWMA, a novel deep learning framework for SWM parcellation, with successful application on datasets across ages and health conditions. SupWMA is lightweight, preserves significant spatial features, and enables discriminative representations of streamlines. Compared to the SOTA methods, SupWMA is the top performer on a ground truth labeled dataset and on three independently acquired testing datasets. SupWMA enables fast and effective SWM parcellation.

---

[7] github.com/SlicerDMRI/whitematteranalysis

## 5. COMPLIANCE WITH ETHICAL STANDARDS

This research study was conducted retrospectively using human subject data made available in open access by Human Connectome Project [22], Adolescent Brain Cognitive Development (ABCD) [23], and Parkinson's Progression Markers Initiative (PPMI) [24]. Ethical approval was not required.

## 6. ACKNOWLEDGEMENTS

We acknowledge the following NIH grants: P41EB015902, R01MH074794, R01MH125860 and R01MH119222. F.Z. also acknowledges a BWH Radiology Research Pilot Grant Award.